\documentclass{article}
\usepackage{ijcai18}
\usepackage{graphicx}
\usepackage{times}
\usepackage{xcolor}
\usepackage{soul}
\usepackage[utf8]{inputenc}
\usepackage[small]{caption}
\usepackage{amsmath}
\usepackage{amssymb}
\usepackage{mathrsfs}
\usepackage{colortbl,booktabs}
\usepackage{multirow}
\usepackage{array}

\title{Sparse Adversarial Perturbations for Videos}

\author{
Xingxing Wei$^1$, 
Jun Zhu$^1$, 
Hang Su$^1$
\\ 
$^1$ Tsinghua National Lab for Information Science and Technology \\
$^1$ State Key Lab of Intelligent Technology and Systems\\
$^1$ Center for Bio-Inspired Computing Research  \\
$^1$ Department of Computer Science and Technology, Tsinghua University\\
\{xwei11, dcszj,  suhangss\}@mail.tsinghua.edu.cn
}

\begin{document}

\maketitle

\begin{abstract}
Although adversarial samples of deep neural networks (DNNs) have been intensively studied on static images,  their extensions in videos are never explored. Compared with images, attacking a video needs to consider not only spatial cues but also temporal cues. Moreover, to improve the imperceptibility as well as reduce the computation cost, perturbations should be added on as fewer frames as possible, i.e., adversarial perturbations are temporally $sparse$. This further motivates the $propagation$ of perturbations, which denotes that perturbations added on the current frame can transfer to the next frames via their temporal interactions. Thus, no (or few) extra perturbations are needed for these frames to misclassify them. To this end, we propose an $l_{2,1}$-norm based optimization algorithm to compute the sparse adversarial perturbations for videos. We choose the action recognition as the targeted task, and networks with a CNN+RNN architecture as threat models to verify our method. Thanks to the propagation, we can compute perturbations on a shortened version video, and then adapt them to the long version video to fool DNNs.  Experimental results on the UCF101 dataset demonstrate that even only one frame in a video is perturbed, the fooling rate can still reach 59.7$\%$.
\end{abstract}

\section{Introduction}

In the past decade, Deep Neural Networks (DNNs) have shown great superiority in computer vision tasks, like image recognition \cite{he2016deep}, image restoration \cite{dong2014learning} and visual tracking \cite{wang2013learning}. Although DNNs obtain the state-of-the-art performance in these tasks, they are known to be vulnerable to adversarial samples~\cite{szegedy2013intriguing}, i.e., the images with visually imperceptible perturbations that can mislead the network to produce wrong predictions. The adversarial samples are  usually calculated by the Fast Gradient Sign Method (FGSM) \cite{goodfellow2014explaining} and optimization-based methods \cite{moosavi2016universal}. One reason for adversarial samples is considered to be that they are fell on some areas in the high-dimensional feature space which are not explored during training. Thus, investigating adversarial samples not only helps understand the working mechanism of deep networks, but also provides opportunities to improve the networks' robustness \cite{xie2017adversarial,dong2017boosting}.

\begin{figure}[t]
\centering\includegraphics[width=0.45\textwidth]{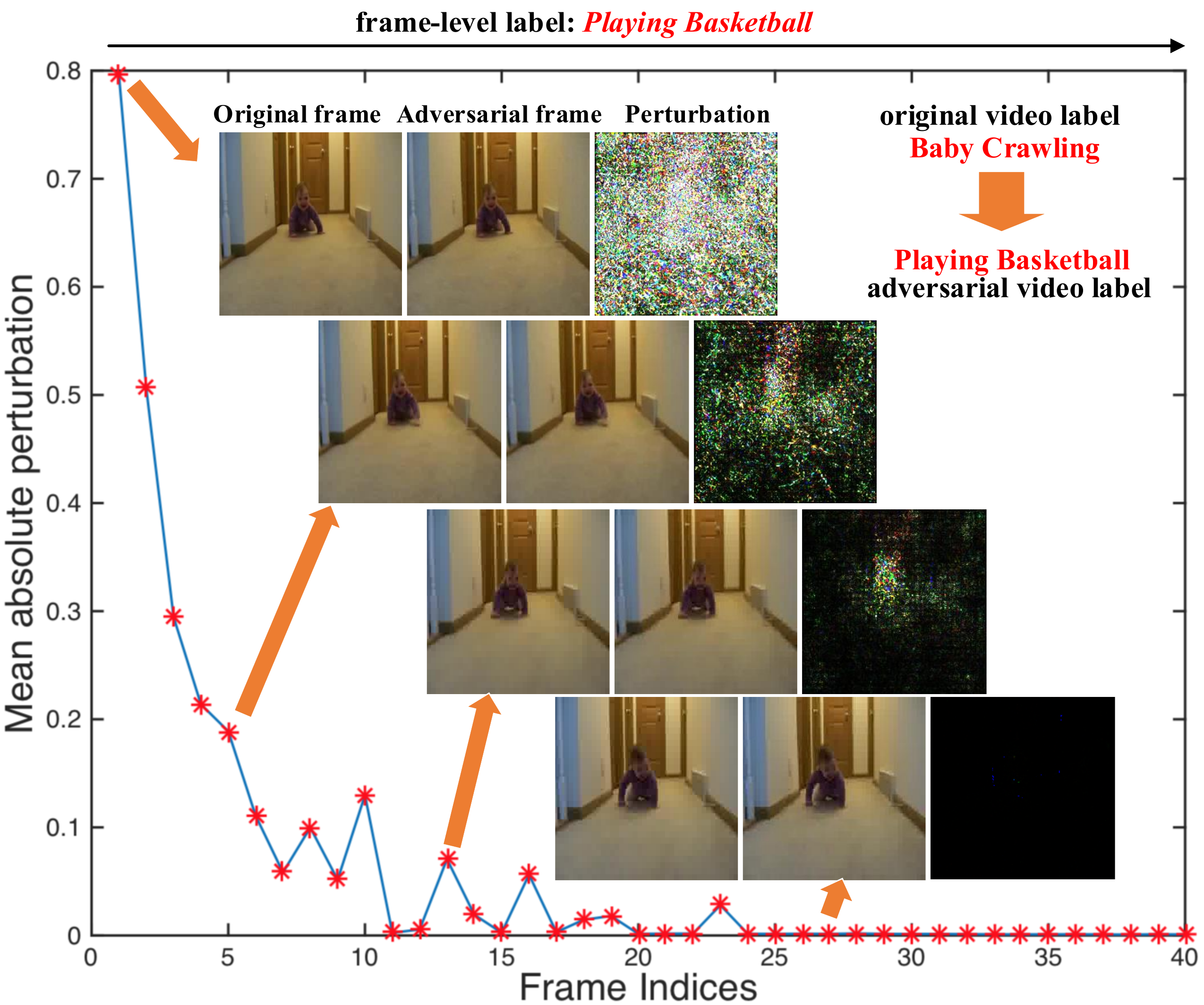}
\vspace{-.2cm}
\caption{An illustration of our output for a video from UCF101 dataset with label of BabyCrawling. The computed perturbations successfully fool DNNs to output label of PlayingBasketball. The Mean Absolute Perturbation (MAP) of each frame is plotted. From the figure, we see that MAP values are significantly reduced along with the varying frames.  In the final 20 frames, they fall into almost zero, showing the $sparsity$ of perturbations. Note that, though these frames have no perturbations, they still make DNNs predict wrong label for each frame (see the top arrow line). That's to say, the perturbations from previous frames $propagate$ here and show their power. Four triples indicate the results from the 1-, 5-, 13-, 27-th frames, respectively. In each triple, we see the adversarial frame is the same to the original frame in the appearance. We better visualize the perturbations with increasing their amplitudes with $\times$ 255.}
\label{fig:figure1}
\vspace{-.4cm}
\end{figure}

Up to now, many studies about adversarial samples have been investigated, such as adversarial perturbations for a single image \cite{moosavi2016deepfool}, universal adversarial perturbations \cite{moosavi2016universal} and adversarial samples for object detection and segmentation \cite{xie2017adversarial}. However, these studies are all based on images, while leaving videos unexplored. Investigating adversarial samples on videos is of both theoretical and practical values, as deep neural networks have been widely applied in video analysis tasks~\cite{donahue2015long,nguyen2015deep,wang2016temporal}. 

Technnically, the main difference between videos and images lies in the temporal structure contained in videos. Therefore, a properly designed attacking method should explore the temporal information to achieve efficiency and effectiveness. We expect that 
the perturbations added on one frame can propagate to other frames via temporal interactions, which will be called the $propagation$ of perturbations. Besides, a video have many frames, computing perturbations for each frame is time-consuming, and actually not necessary. Whether it is possible that perturbations are added on only few frames, and then are propagated to other frames to misclassify the whole video. In this way, the generated adversarial videos also have high imperceptibility and are hard to be detected. Because perturbations are added on sparse frames rather than the whole video, we call it the $sparsity$ of perturbations. Actually, the $propagation$ and $sparsity$ interact with each other, $propagation$ helps boost the $sparsity$, meanwhile the $sparsity$ constraint will lead to better $propagation$.
\begin{figure}[t]
\centering\includegraphics[width=0.35\textwidth]{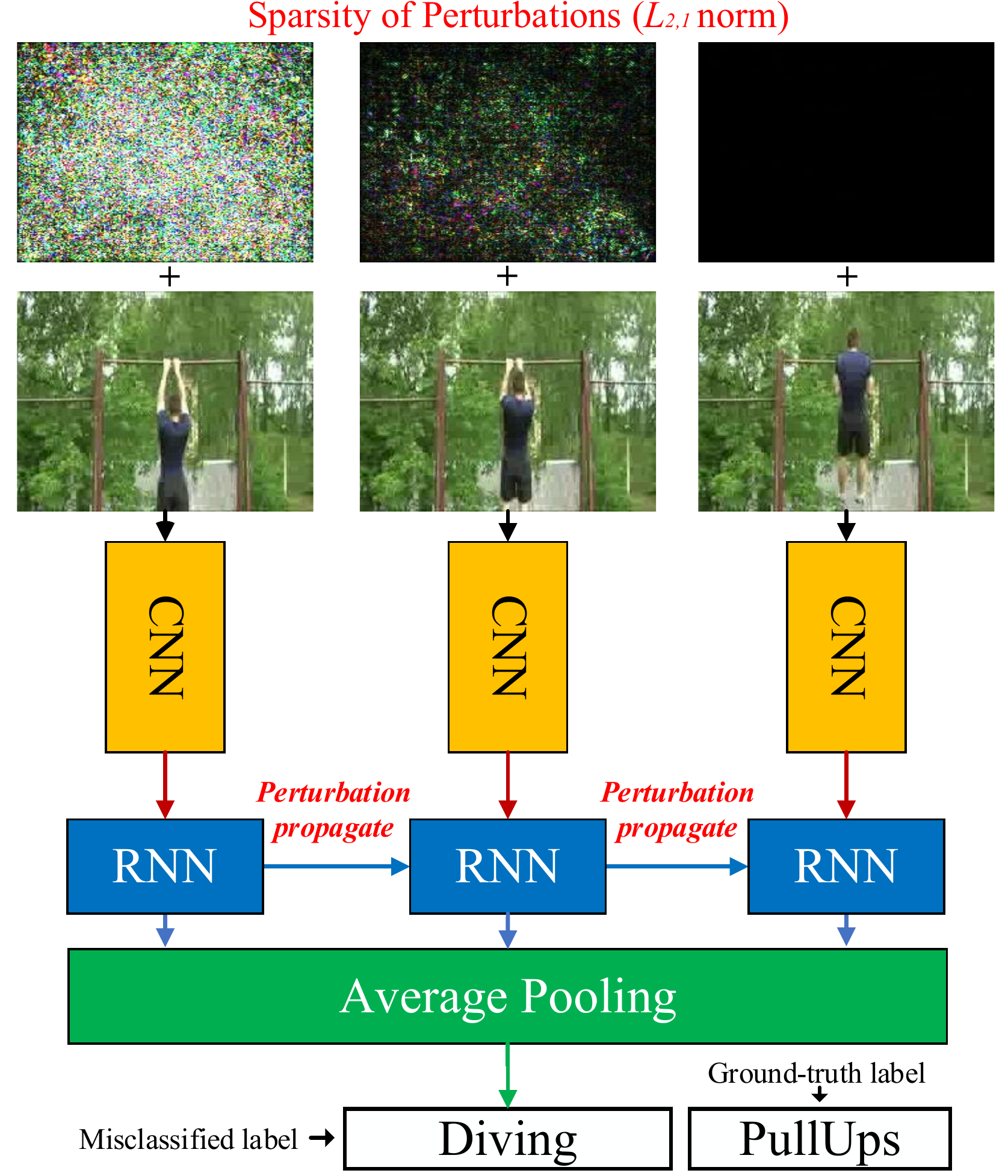}
\vspace{-.2cm}
\caption{The illustration of our method. An $l_{2,1}$ norm is used during the optimization, which enforces the sparsity of computed perturbations.  Within the CNN+RNN architecture (action recognition network), perturbations are encoded after the CNN, and then propagate to the next frames via RNN, finally resulting in the misclassified label for the whole video. Thanks to the propagation of perturbations, perturbations added on the final frames fall into zeros. }
\label{fig:figure2}
\vspace{-.4cm}
\end{figure}

For these reasons, in this paper, we aim to attack the video action recognition task \cite{poppe2010survey}, where the temporal cue is a key component for the predicted label. This is naturally suitable to explore the temporal adversarial perturbations. For the threat model, we choose the networks with a CNN+RNN architecture, which is widely used in action recognition, such as Long-term Recurrent Convolutional Network (LRCN) \cite{donahue2015long} or network in \cite{yue2015beyond}. To achieve $sparsity$. we apply an $l_{2,1}$-norm regularization on perturbations during the optimization. The $l_{2,1}$-norm uses the $l_1$ norm across frames, and thus, enforces to select few key frames to add perturbations.  As for $propagation$, we find perturbations show good propagation under the $l_{2,1}$ constraint within the recurrent neural network (such as Vanilla RNN, LSTM and GRU) because of the interaction with $sparsity$. Another advantage of the propagation is that we can compute perturbations on a shortened version video, and then adapt them to the long version video to fool DNNs, which provides a more efficient method to attack videos. The illustrations of our output and method are given in Fig.~\ref{fig:figure1} and Fig.~\ref{fig:figure2}, respectively.

In summary, this paper has the following contributions:
\begin{itemize}
\item To our knowledge, we are the first to explore adversarial samples in videos. Considering the specific $sparsity$ and $propagation$ of video adversarial perturbations,  we propose an $l_{2,1}$-norm regularization based optimization algorithm. We verify our method and evaluate its transferability on the UCF101 dataset.

\item We give a comprehensive evaluation of the $sparsity$ and $propagation$ of perturbations, and furthermore, propose the propagation-based method for adversarial videos, i.e., computing perturbations on a shortened version video, and then adapt them to the long version video. We also find that LSTM and GRU are easier to be attacked than Vanilla RNN, because LSTM and GRU can represent long memory, which is favor to the perturbation propagation (see experiments). 

\end{itemize}

The rest of this paper is organized as follows. In Section 2, we briefly review the related work. We present our  algorithm in Section 3. Section 4 reports all experimental results. Finally, we summarize the conclusions in Section 5.

\section{Related Work}
The related work comes from two aspects: action recognition with deep learning and adversarial attack.
\subsection{Action Recognition with Deep Learning}
Action recognition is a core task in computer vision, where its goal is to predict a video-level label when given a video clip \cite{poppe2010survey}.
With the rise of deep convolutional neural networks (CNNs) achieving state-of-the-art performance on image recognition, many works have looked into designing effective deep CNNs for action recognition. For instance, various approaches of fusing CNN features computed on RGB frames over the temporal dimension are explored on the
Sport1M dataset \cite{karpathy2014large}. To integrate the temporal information, CNN+LSTM based models, which use a CNN to extract frame features and an LSTM to integrate features over time, are also presented to recognize activities in videos \cite{donahue2015long,nguyen2015deep}. Optical flow is also useful to encode the temporal cue. For this, two stream CNNs with one stream of static images and the other stream of optical flows are proposed to fuse the information of object appearance and short-term motions \cite{simonyan2014two}. Temporal Segment Networks sample frames and optical flow on different time segments to extract information for activity recognition \cite{wang2016temporal}.  In our paper, to better explore how the perturbations change along with the time, we choose the networks with a CNN+RNN architecture as the threat model.
\subsection{Adversarial Attack}

Generating adversarial examples for classification has been extensively studied in many different ways recently. \citeauthor{szegedy2013intriguing}~[2013] first show that adversarial examples, computed by adding visually imperceptible perturbations to the original images, make CNNs predict a wrong label with high confidence. \citeauthor{goodfellow2014explaining}~[2014] propose a simple and fast gradient sign method to generate adversarial examples based on the linear nature of CNNs. \citeauthor{moosavi2016deepfool}~[2016b] propose a simple algorithm to compute the minimal adversarial perturbation by assuming that the loss function can be linearized around the current data point at each iteration. \citeauthor{moosavi2016universal}~[2016a] show the existence of universal (image-agnostic) adversarial perturbations. \citeauthor{baluja2017adversarial}~[2017] train a network to generate adversarial examples for a particular target model (without using gradients). \citeauthor{kurakin2016adversarial}~[2016] show that the adversarial examples for machine learning systems also exist in the physical world. \citeauthor{liu2016delving}~[2016] study the transferability of both non-targeted and targeted adversarial examples, and proposed an ensemble-based approaches to generate adversarial examples with stronger transferability.  The above papers are all based on images, while we focus on video adversarial samples, which have new challenges. 

\section{Methodology}

In this section, we introduce the proposed $l_{2,1}$-norm based algorithm for  video adversarial samples. Our method is an optimization-based approach. 

Let $\mathbf{X}\in\mathbb{R}^{T \times W \times H \times C}$ denote a clean video, and $\mathbf{\hat{X}}$ denote its adversarial video, where $T$ is the number of frames, $W, H, C$ are the width, height, and channel for a specific frame, respectively. $\mathbf{E}=\mathbf{\hat{X}}-\mathbf{X}$ is the adversarial perturbations. To generate non-targeted adversarial examples, we approximate the solution to the following objective function:
\begin{equation}\label{eq:eq1}
\arg\min_{\mathbf{E}} \lambda ||\mathbf{E}||_{p} -\ell(\mathbf{1}_y, \emph{J}_\theta(\mathbf{\hat{X}})),
\end{equation}
where $\ell(,\cdot ,)$ is the loss function to measure the difference between the prediction and the ground truth label.  In this paper, we choose the widely used cross-entropy function $\ell(u,v) = \log (1-u \cdot v)$, which is shown to be effective~\cite{carlini2017towards}.  $\emph{J}_\theta(\cdot)$ is the threat model with parameters $\theta$. $\mathbf{1}_y$ is the one-hot encoding of the ground truth label $y$.  $||\mathbf{E}||_p$ is the $\ell_p$ norm of $\mathbf{E}$, which is a metric to quantify the magnitude of the perturbation. $\lambda$ is a constant to balance the two terms in the objective. 

To obtain a universal adversarial perturbation across videos, we solve the following problem:
\begin{equation}\label{eq:eq2}
\arg\min_{\mathbf{E}} \lambda ||\mathbf{E}||_{p} -\frac{1}{N}\sum_{i=1}^N\ell(\mathbf{1}_{y_i}, \emph{J}_\theta(\mathbf{\hat{X}}_i)),
\end{equation}
where $N$ is total number of training videos, and $\mathbf{\hat{X}}_i$ is the $i$-th adversarial video. 

To better control the sparsity and study the perturbation propagation across frames, we add a temporal mask on the video to enforce some frames having no perturbations. The problem is modified as follows:
\begin{equation}\label{eq:eq3}
\arg\min_{\mathbf{E}} \lambda ||\mathbf{M\cdot E}||_{p} -\frac{1}{N}\sum_{i=1}^N\ell(\mathbf{1}_{y_i}, \emph{J}_\theta(\mathbf{X}_i+\mathbf{M\cdot E})),
\end{equation}
where $\mathbf{M}\in \{\mathbf{0,1}\}^{T \times W \times H \times C}$ is the temporal mask.  We let $\Theta=\{1,2,..., T\}$ be the set of frame indices, $\Phi$ is a subset within $\Theta$ having \emph{K} elements,  and $\Psi=\Theta-\Phi$. If $t \in \Phi$,  we set $M_t=0$, and if $t \in \Psi, M_t=1$, where $M_t\in\{\mathbf{0,1}\}^{W \times H \times C}$ is the \emph{t}-th frame in $\mathbf{M}$. In this way, we enforce the computed perturbations to be added only  on the selected video frames. We here regard $S=\frac{K}{T}$ as the sparsity. 

If the goal is to generate targeted adversarial examples (i.e., the misclassified label is set to the pre-fixed label, which is called target label),  the problem can be modified as follows:
\begin{equation}\label{eq:eq4}
\arg\min_{\mathbf{E}} \lambda ||\mathbf{M\cdot E}||_{p} +\frac{1}{N}\sum_{i=1}^N\ell(\mathbf{1}_{y_i^*}, \emph{J}_\theta(\mathbf{X}_i+\mathbf{M\cdot E})),
\end{equation}
where $y_i^*$ is the targeted label. Eq.(\ref{eq:eq4}) outputs the perturbations to make $\emph{J}_\theta(\cdot)$ predict $y_i^*$ with a high probability.

\subsubsection{Perturbation Regularization}
The $l_p$-norm in problem~(\ref{eq:eq1},\ref{eq:eq2},\ref{eq:eq3},\ref{eq:eq4})  is a metric to quantify the magnitude of the perturbation. As mentioned before, we hope that the perturbations are added on as fewer frames as possible. Therefore, we choose $l_{2,1}$ norm to meet this goal, where is widely used in sparse coding methods~\cite{wright2009robust,yang2010image}. $||\mathbf{E}||_{2,1}=\sum_t^T||E_t||_2$, where $E_t\in\mathbb{R}^{W \times H \times C}$ is the \emph{t}-th frame in $\mathbf{E}$. $l_{2,1}$ norm apply the $l_1$ norm across the frames, and thus, can ensure the sparsity of generated perturbations. In the experiment, we also show the results using $l_{2}$ norm, as the comparison with the $l_{2,1}$ norm.

\subsubsection{Threat Model}
In action recognition, the current state-of-the-art approach is the two-stream model \cite{donahue2015long}, i.e., one stream is to capture the RGB frames, and another stream is to capture the optical flow images (motion information) between two adjacent RGB frames. The outputs from these two streams are fused to predict the final label with various kinds of fusion methods. These two streams usually have the same network architecture, where one choice is CNN+Pooling, and another is CNN+RNN architecture. Compared with CNN+Pooling, CNN+RNN can encode the temporal information. In our paper, we regard the networks with CNN+RNN architecture as the threat model $\emph{J}_\theta(\cdot)$. The results of attacking CNN+Pooling also are reported for comparisons. We give the illustration of the CNN+RNN model in Fig.~\ref{fig:figure2}. Note that, the CNN and RNN in the figure are the general terms for the spatial and temporal networks, respectively. CNN can be specified as ResNet, Inception V3, etc, and RNN as LSTM, GRU, etc.
\begin{figure*}[t]
\centering\includegraphics[width=0.92\textwidth]{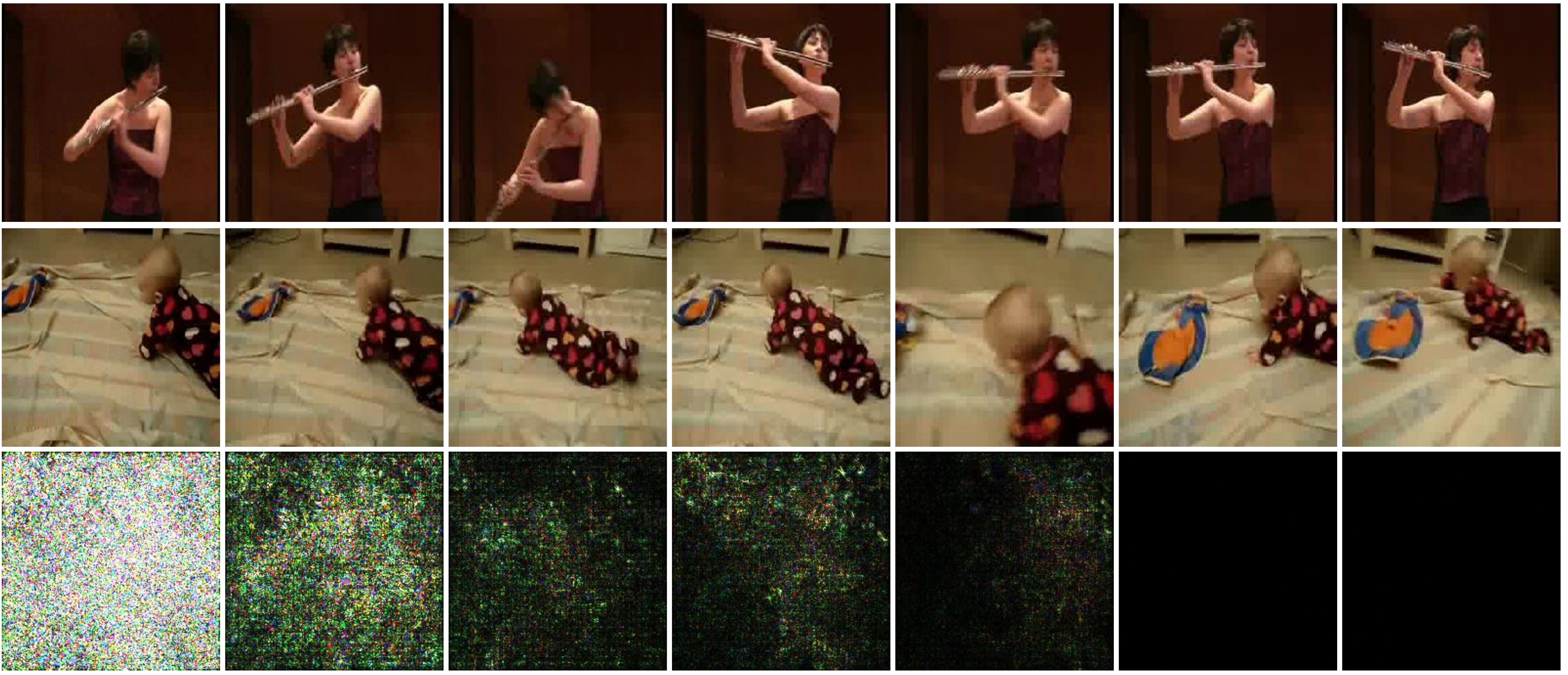}
\vspace{-.2cm}
\caption{The computed adversarial videos (top two rows) and their corresponding perturbations (bottom row) using Eq.(\ref{eq:eq2}) with $l_{2,1}$ norm. We better visualize the perturbations with increasing their amplitudes with $\times$ 255. For more discussions, please see the texts.}
\label{fig:figure9}
\vspace{-.2cm}
\end{figure*}

\subsubsection{Training}
Problems~(\ref{eq:eq1},\ref{eq:eq2},\ref{eq:eq3},\ref{eq:eq4}) are easy to solve. Any Stochastic Gradient Descent (SGD) algorithm can solve them. Here, we use the Adam \cite{kingma2014adam} algorithm to get the results. Because $l_{2,1}$ norm is used, initializing the perturbations with zeros will lead to NaN values. We instead initialize them using a small value. In the experiments, we use 0.0001. After some iterations, the perturbations will  converge to a sparse result. $\lambda$ in problem~(\ref{eq:eq1},\ref{eq:eq2},\ref{eq:eq3},\ref{eq:eq4}) is set to a constant, which is tuned in the training set. Temporal mask $\mathbf{M}$ is predefined according to the needed sparsity. We investigate some choices, and give the corresponding discussions about its impact to the proposed method (see experiments).

\section{Experiments}
In this section, we give the experiments from three aspects.
\subsection{Datasets and Metrics}

{\bf Datasets:} We choose the widely used dataset in action recognition: $\mathbf{UCF101}$ dataset \cite{soomro2012ucf101}. It contains 13,320 videos with 101 action classes covering a broad set of activities such as sports, musical instruments, body-motion, human-human interaction, human-object interaction. The dataset splits more than 8000 videos in the training set, and more than 3000 videos in the testing set.  

Because there are no other existing methods for video adversarial samples, we can only compare with the methods based on images, i.e., computing perturbations for each frame \cite{moosavi2016deepfool} in a video. This setting is coincident with the outputs using Eq.(\ref{eq:eq1}) with $l_2$ norm, which are reported as the comparisons. 


\noindent {\bf Metrics:} We use three metrics to evaluate various aspects.

$\mathbf{Fooling}$  $\mathbf{ratio}$ ($\mathbf{F}$): is defined as the percentage of adversarial videos that are successfully misclassified \cite{moosavi2016universal}.

$\mathbf{Perceptibility}$ ($\mathbf{P}$): denotes the perceptibility score of the adversarial perturbation $\mathbf{r}$. We here use the Mean Absolute Perturbation (MAP): $P=\frac{1}{N}\sum_i|\mathbf{r}_i|$, where N is the number of pixels, and $\mathbf{r}_i$ is the intensity vector (3-dimensional in the RGB color space).

$\mathbf{Sparsity}$ ($\mathbf{S}$): denotes the proportion of frames with no perturbations (clean frames) versus all the frames in a specific video to fool DNNs. $S=\frac{K}{T}$, where $K$ is the number of clean frames, and $T$ is the total number of frames in a video.

\begin{figure*}[t]
\centering\includegraphics[width=0.245\textwidth]{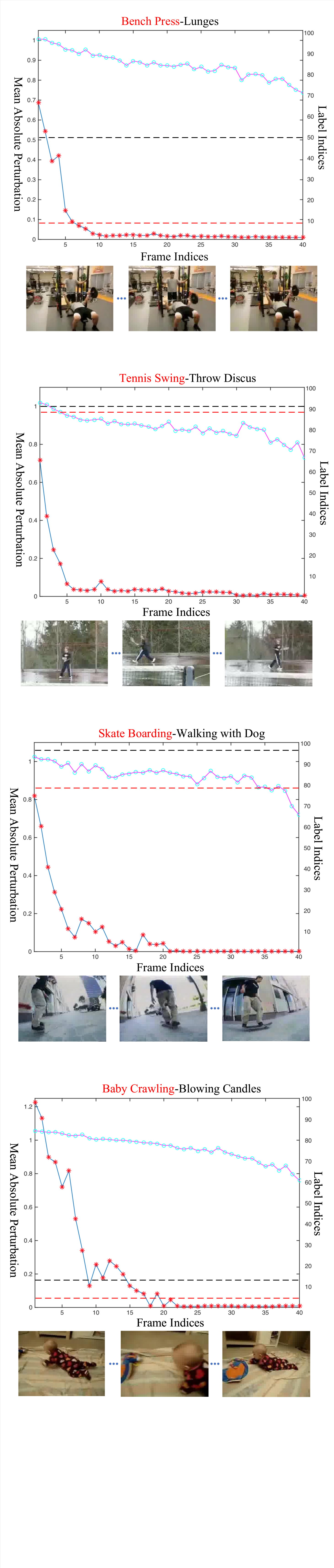}
\centering\includegraphics[width=0.245\textwidth]{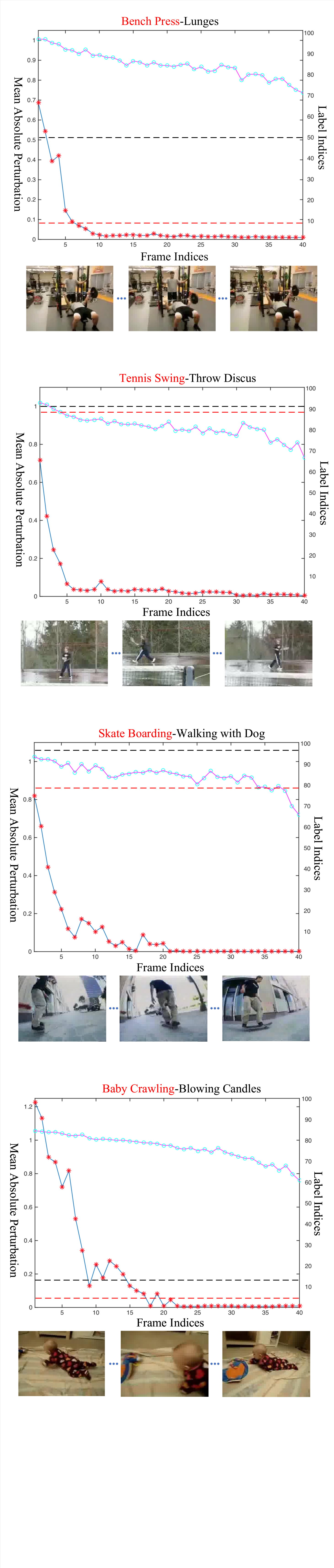}
\centering\includegraphics[width=0.245\textwidth]{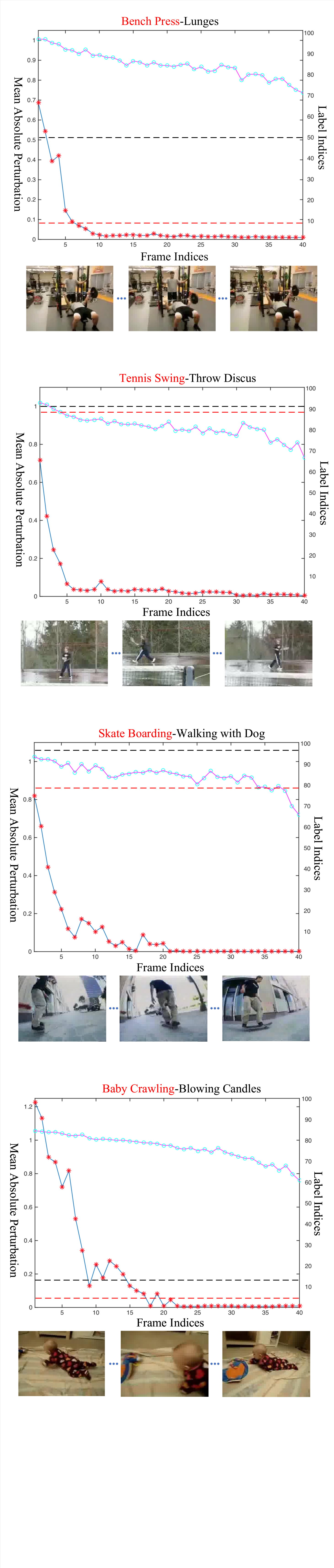}
\centering\includegraphics[width=0.245\textwidth]{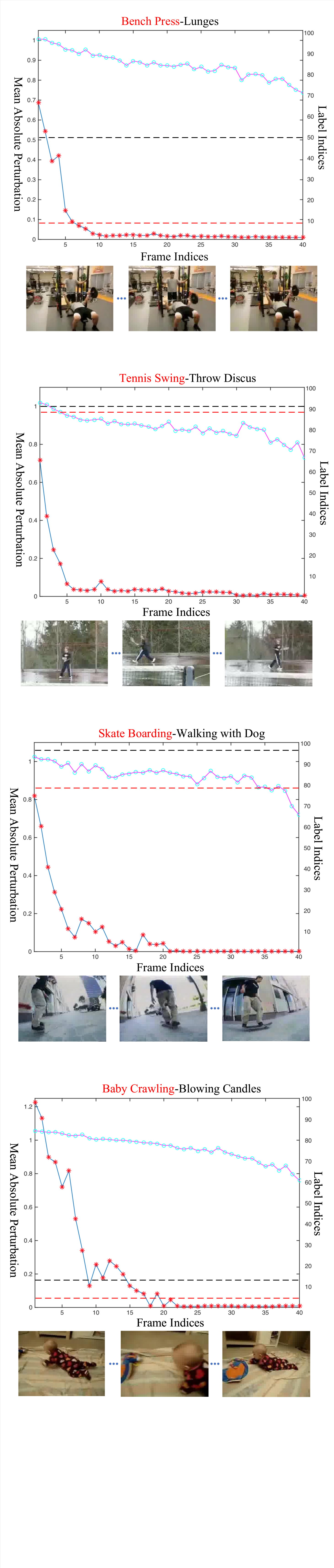}
\vspace{-.2cm}
\caption{Four examples for showing perturbation propagation on UCF101 dataset. The $x$-axis denotes the frame indices in a video. The left $y$-axis denotes the Mean Absolute Perturbation (MAP) value of each frame's perturbations, and the right $y$-axis is the label indices. The blue line with stars is the curve of MAP values with $l_{2,1}$ norm, and magenta line with circles is the result with $l_2$ norm. The red dotted line is the predicted frame-level label indices for the clean video, and black dotted line is the predicted frame-level label indices for the adversarial video, both by the action recognition networks (the video-level labels are listed in the top of each figure with the same color). In the bottom of each figure, we give the corresponding video frames. For detailed discussions, please see the texts.}
\label{fig:figure3}
\vspace{-.2cm}
\end{figure*}
\subsection{Perturbation Propagation}
In this section, we give the experimental results about the perturbation propagation.

\begin{figure*}[t]
\centering\includegraphics[width=0.245\textwidth]{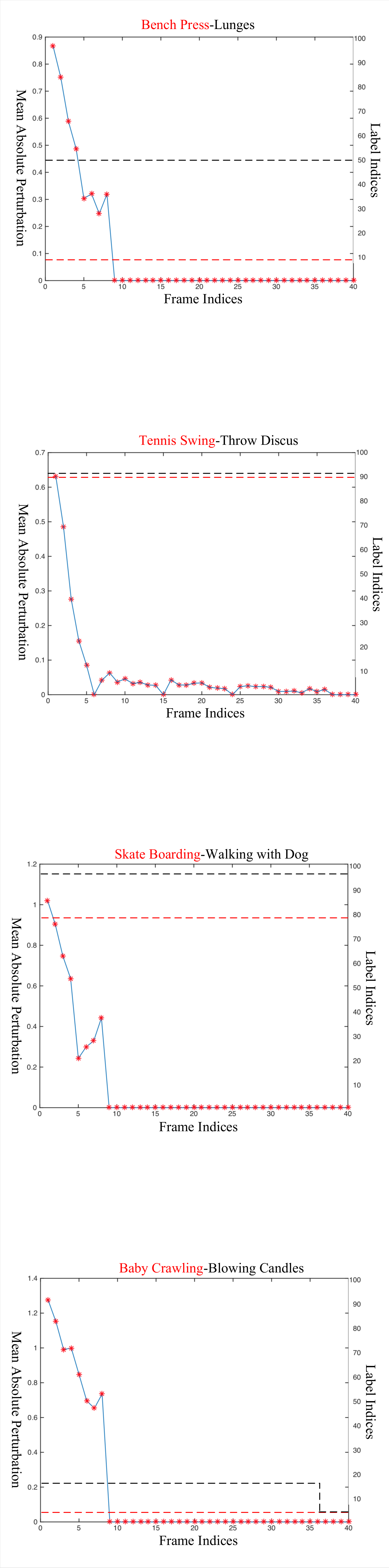}
\centering\includegraphics[width=0.245\textwidth]{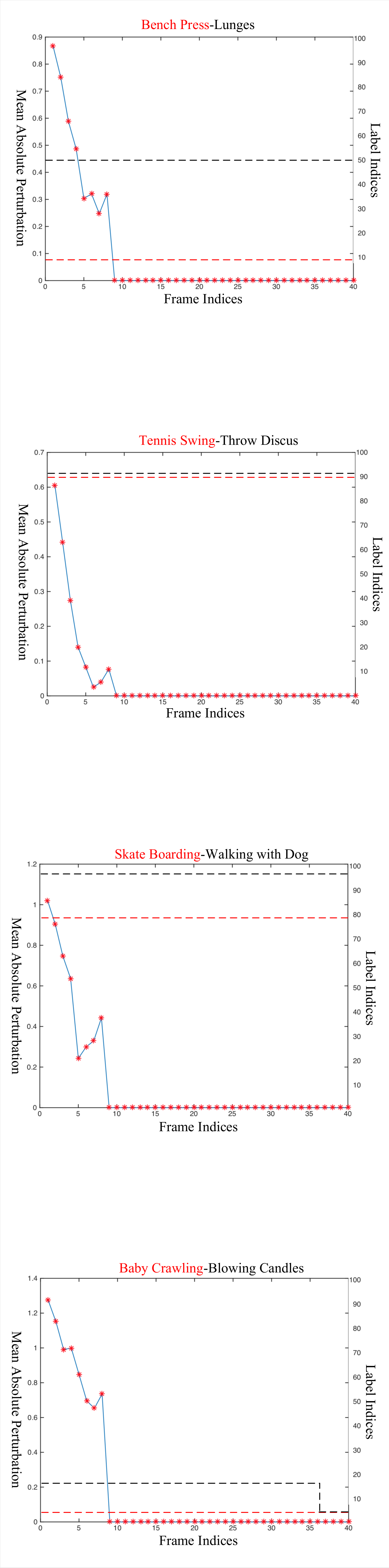}
\centering\includegraphics[width=0.245\textwidth]{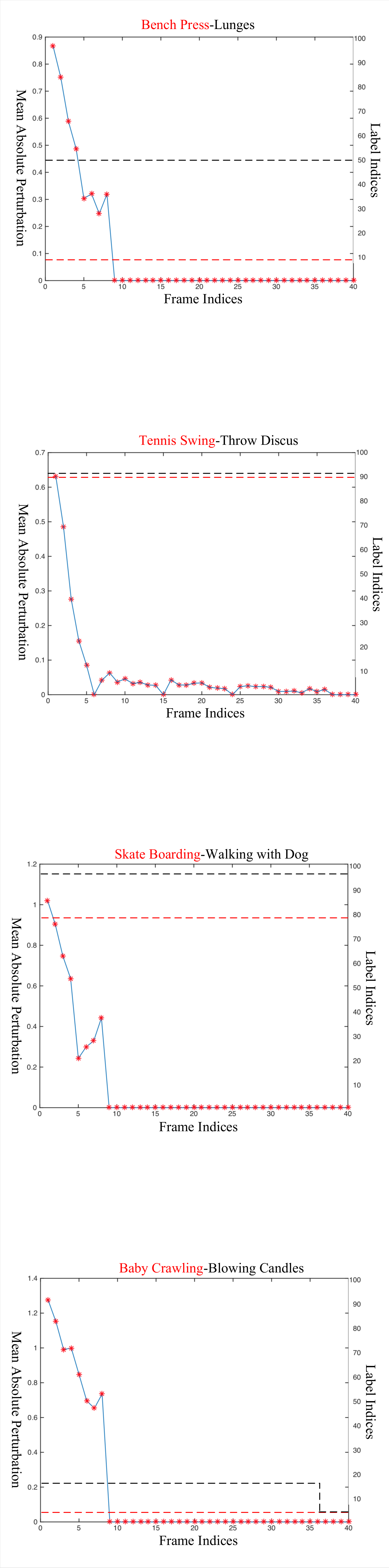}
\centering\includegraphics[width=0.245\textwidth]{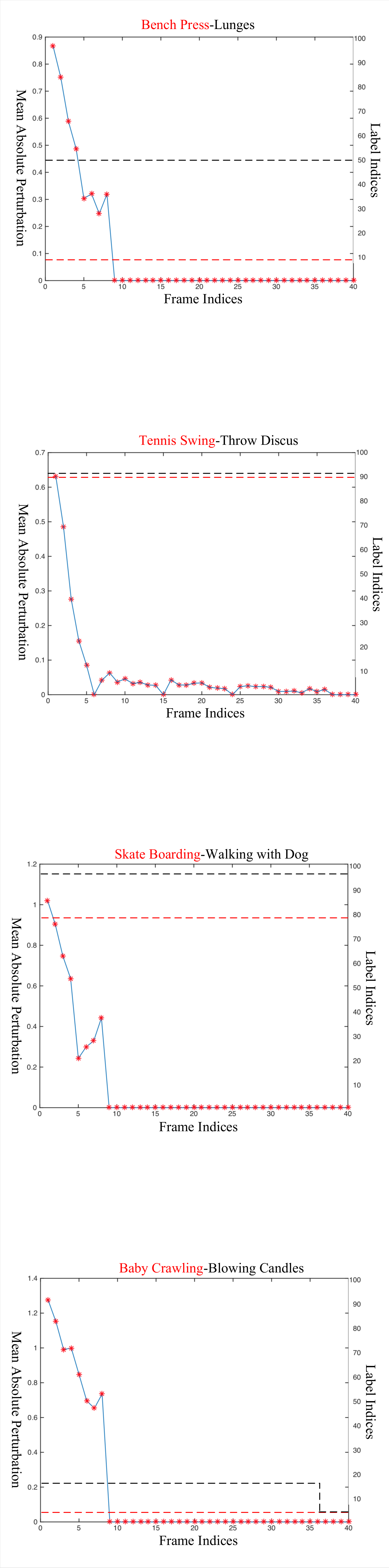}
\vspace{-.2cm}
\caption{Four examples of showing perturbation propagation on UCF101 dataset. The difference with Fig.~\ref{fig:figure3} lies in the integration of temporal mask proposed in Eq.(\ref{eq:eq3}). For detailed discussions, please see the texts.}
\label{fig:figure4}
\vspace{-.2cm}
\end{figure*}

\subsubsection{Visualization for Perturbations}
We firstly give the visualization of perturbations computed using Eq.(\ref{eq:eq2}) with $l_{2,1}$ norm, which are universal perturbations across videos. In Fig.~\ref{fig:figure9},  we see that the adversarial videos are not distorted by the perturbations, and are imperceptible to human eyes. Furthermore, the perturbations show the sparse property (black means no perturbations), i.e., they are reduced across frames along with the time, which is owing to the used $l_{2,1}$ norm. In the next section, we will discuss the propagation of perturbations, inspired by these sparse results.

\subsubsection{Perturbation Propagation}
To show the perturbation propagation, we give four examples outputted by Eq.(\ref{eq:eq1}) with $l_{2,1}$ norm in Fig.~\ref{fig:figure3} (see the blue line with stars), where we see the computed perturbations successfully fool the action recognition networks (for example, in the first case, a clean video with label of Bench Press is identified as Lunges after adding perturbations). Correspondingly, the original frame-level labels (red dotted line) are also misclassified as wrong labels (black dotted line). By contrast, the Mean Absolute Perturbation (MAP) value of each frame is reduced significantly along with the time. In the last few frames, they fall into almost zeros. That's to say, although few perturbations are added on these frames, the perturbations from the previous frames propagate here, and help fool the DNNs. As a comparison, we also list the results of Eq.(\ref{eq:eq1}) with $l_{2}$ norm in Fig.~\ref{fig:figure3} (see the magenta line with circles). In this figure, the MAP value is also reduced across frames, which further demonstrates the perturbation propagation. The difference is, the output of $l_{2,1}$ norm is sparse, which reveals that the frames ranking behind the video line actually need few (even no) perturbations to fool DNNs with the help of propagation. But $l_2$ norm cannot show this property.

Inspired by the sparsity of $l_{2,1}$ norm,  we directly enforce perturbations not to be added on the frames ranking behind the video line.  To this end, we add the temporal mask during the optimization process using Eq.\ref{eq:eq3}. Here we only select the top 8 frames to compute their perturbations, and let the other frames be clean. The experimental results on the same videos are listed in Fig.~\ref{fig:figure4}. We find that the frames are still predicted as wrong labels. Furthermore, the MAP values of these frames also show a decreasing trend. It is further demonstrated the propagation of perturbations. Otherwise, these clean frames cannot be predicted as wrong labels. Note that in the forth case in Fig.~\ref{fig:figure4},  the final 4 frames have correct labels, which shows perturbations will reduce its effect along with the time, and cannot propagate forever.

\begin{table}[htbp]\caption{The results of fooling rates versus different sparsities.} \centering
\begin{tabular}{|p{1.25cm}<{\centering}|p{1.25cm}<{\centering}|p{1.25cm}<{\centering}|p{1.25cm}<{\centering}|p{1.25cm}<{\centering}|}
\hline
\textbf{S}          &0$\%$(40)  &80$\%$(8) &90$\%$(4) &97.5$\%$(1)   \\
\hline
$\mathbf{F}$              &100$\%$            &100$\%$     &91.8$\%$    &59.7$\%$  \\
\hline
$\mathbf{P}$               &0.0698            &0.6145     &1.0504    &1.9319  \\
\hline
\end{tabular}\label{tab:tab6}
\end{table}

\begin{figure}[t]
\centering\includegraphics[width=0.48\textwidth]{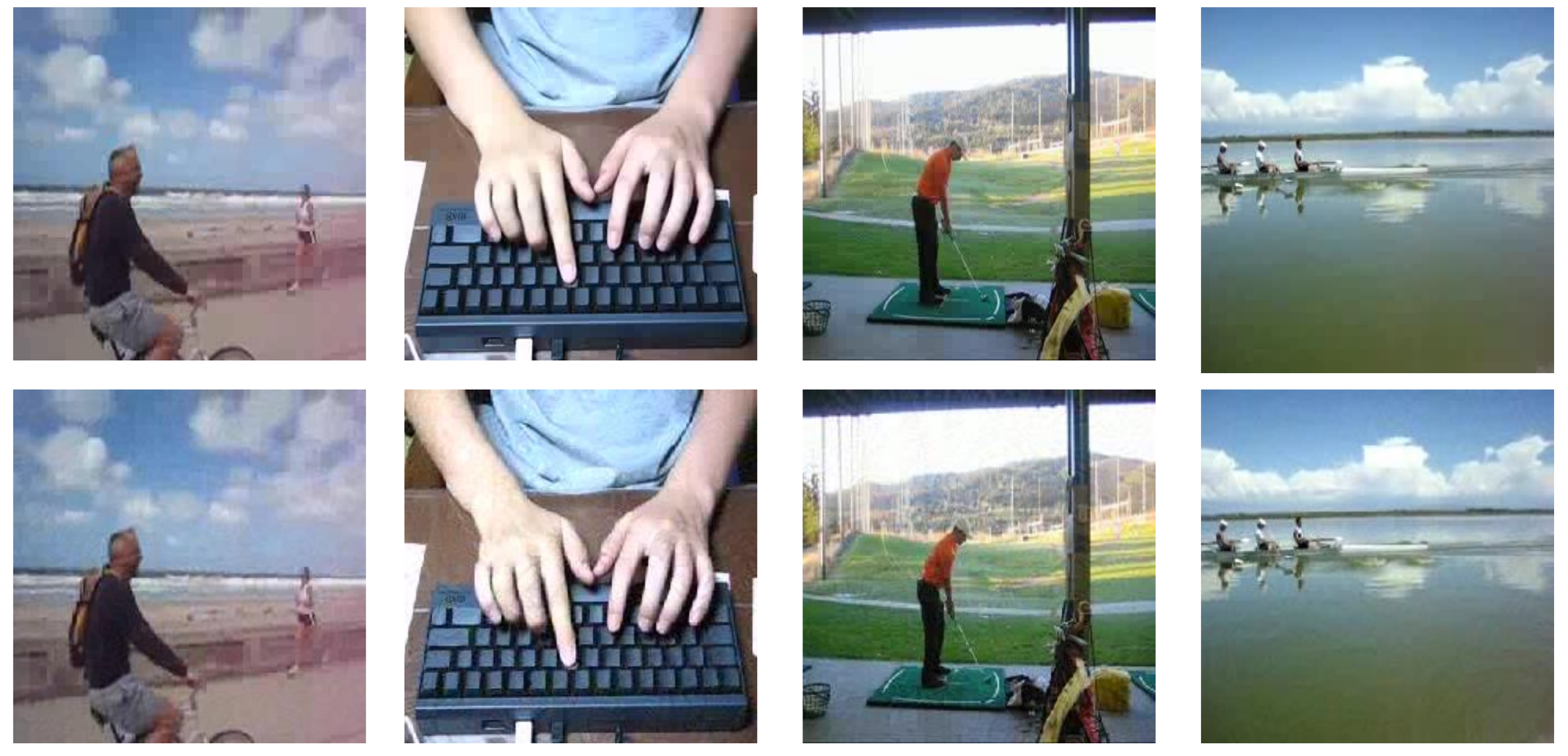}
\vspace{-.5cm}
\caption{Four examples of the polluted top one frame in $S=97.5\%$. The top row is the original clean frames, and the bottom row is the adversarial frames.}
\label{fig:figure5}
\vspace{-.2cm}
\end{figure}

We now gradually enlarge the sparsity $S$ in Eq.(\ref{eq:eq3}), and observe the change of Fooling ratio $F$ in the testing set on UCF101 dataset. High sparsity $S$ means more clean frames, and less adversarial frames in the video. We give the quantitative results of fooling rates versus different sparsities in Table. \ref{tab:tab6}. In the table, we list four sparsities ($S$) and their corresponding Fooling rates ($F$) as well as perceptibility scores ($P$). Taking $90\%(4)$ as an example,  $90\%=1-\frac{4}{40}$, where 4 is top four polluted frames, and 40 is the total number of frames in the video. The results in Table \ref{tab:tab6} show that even only one frame is polluted ($S=97.5\%$), the Fooling rate can also reach $59.7\%$.  To achieve the $100\%$ fooling rate, the least number of polluted frames is 8 ($S=80\%$) on the used dataset. We also see that the perceptibility score is gradually increasing with the rise of sparsity score, and reach the top in $S=97.5\%$.  This is reasonable because large perturbations can spread to more frames. The polluted top one frames in $S=97.5\%$ and their corresponding clean frames are illustrated in Fig.~\ref{fig:figure5}, where we see that despite the largest $P =1.9319$, the adversarial frames are the same to the clean frames, which are not perceptible to human eyes.

\subsubsection{Adversarial Video based on Propagation}

Thanks to the perturbation propagation, we don't need to compute perturbations based on the whole video. Instead, we can
compute perturbations on a shorten version video, and then adapt them to the long version video. In this way, the computation cost is reduced significantly. We report the time of computing perturbations for various frames in Table \ref{tab:tab7}, where we see the computing time is linearly reduced with the rise of sparsity, showing that computing perturbations on a shorten version video can reduce computation cost.

\begin{figure}[t]
\centering\includegraphics[width=0.35\textwidth]{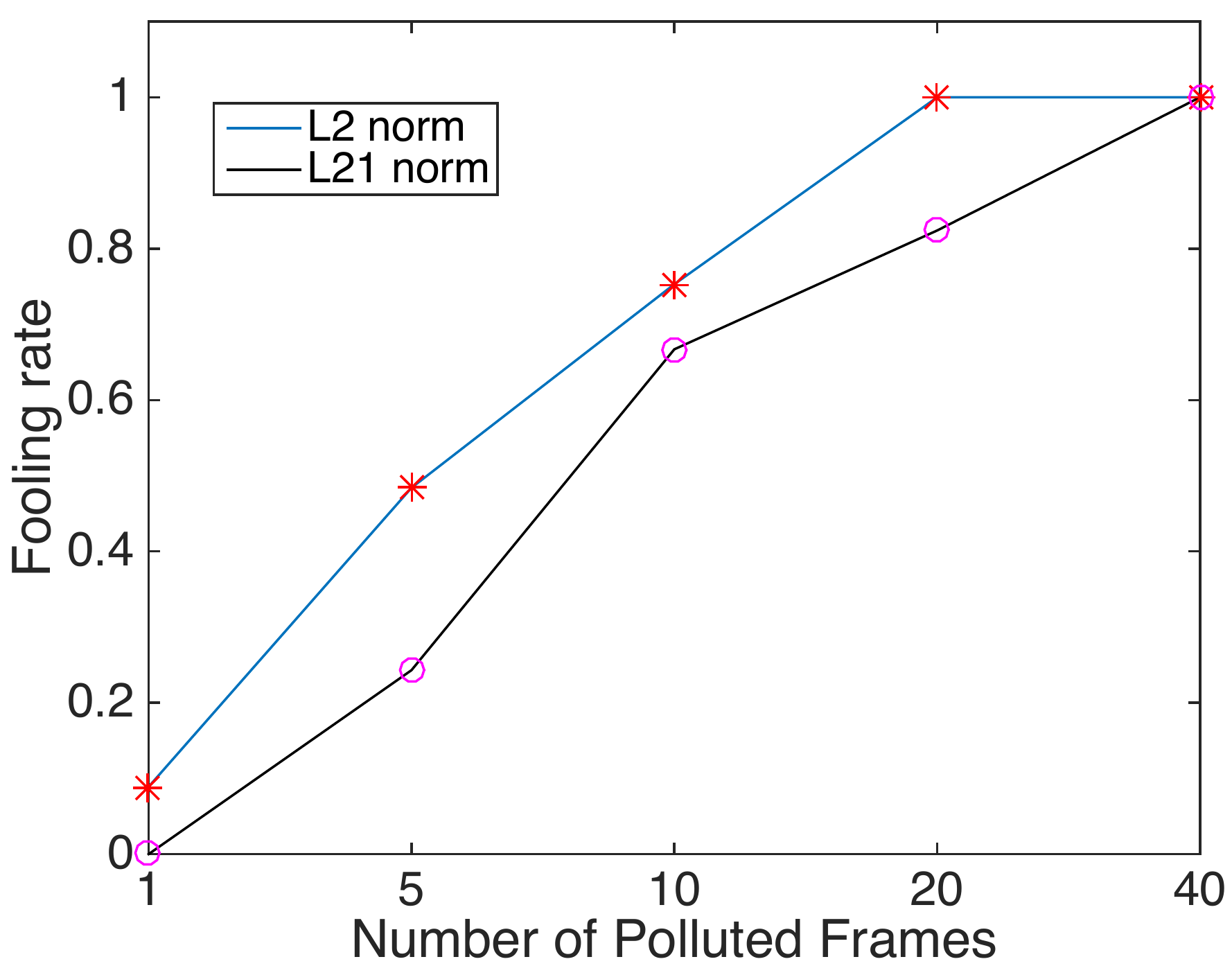}
\vspace{-.2cm}
\caption{Comparisons between $l_2$ and $l_{2,1}$ norm versus Fooling rate on UCF101 dataset. We here report the results when $N=1,5,10,20,40$, respectively. The total number of frames is 40.}
\label{fig:figure6}
\vspace{-.2cm}
\end{figure}

Specifically, to fool the action recognition network for a given video, we first choose the top $N$ frames $\{F_1,..., F_N\}$ from the original video , and then use Eq.(\ref{eq:eq1})  (for a single video) or Eq.(\ref{eq:eq2}) (for getting universal perturbations) with $l_2$ norm to compute their adversarial frames $\{\hat{F}_1,..., \hat{F}_N\}$. Finally, we replace $\{F_1,..., F_N\}$ with $\{\hat{F}_1,..., \hat{F}_N\}$ in the original video.  This modified video is then input to the action recognition networks. Note that, we here don't use the $l_{2,1}$ norm. Because the $l_{2,1}$ norm will result in the sparse perturbations during these $N$ frames, which are not good for further propagation to the rest clean frames. We plot the comparisons between $l_2$ and $l_{2,1}$ norm in this setting in Fig.~\ref{fig:figure6}. In this figure, we see the performance of $l_2$ norm is advantageous to $l_{2,1}$ norm. In the next section, we will give the detailed evaluations and discussions of this method. In default, we set $N=20$ and use $l_2$ norm in the following experiments.

\begin{table}[htbp]\caption{Time for computing perturbations in one iteration.} \centering
\begin{tabular}{|p{1cm}<{\centering}|p{1cm}<{\centering}|p{1cm}<{\centering}|p{1cm}<{\centering}|p{1cm}<{\centering}|p{1cm}<{\centering}|}
\hline
\textbf{S}          &0$\%$  &50$\%$ &75$\%$ &87.5$\%$ &97.5$\%$   \\
\hline
$\mathbf{Time}$        &2.853s       &1.367s      &0.612s     &0.346s    &0.0947s  \\
\hline
\end{tabular}\label{tab:tab7}
\vspace{-.4cm}
\end{table}

\subsection{Performance and Transferability}
In this section, we evaluate the performance and transferability of the propagation based method.
\subsubsection{Transferability across Models}
We firstly evaluate the transferability of computed perturbations. Because the transferability of CNN networks has been studied in many literatures, we here mainly explore the RNN networks, including Vanilla RNN, LSTM, and GRU. Besides, the results of CNN + Average Pooling (removing the RNN layer in Fig.~\ref{fig:figure2}) are also reported. The Fooling rates in different settings are given in Table \ref{tab:tab3}, where we use the networks in  rows to generate perturbations, and networks in columns evaluate the transferability. Form the table, we draw the following conclusions: 
1. The diagonals have largest values. It is reasonable because they perform the white-box attack in this setting. 2. In the off-diagonals, the values are all above $65\%$, which shows perturbations in videos have good transferability, especially in the RNN models. 3. In the off-diagonals, the $Pooling$ column has the poor performance. Pooling method has no memory like LSTM or GRU, and thus, the perturbations cannot propagate to other frames, resulting in the poor performance. 4. By contrast, the $GRU$ and $LSTM$ columns have better performance than VanillaRNN. As we known, GRU and LSTM can represent long memory, this demonstrates long memory is favor to the propagation of perturbations, and thus GRU and LSTM are easier to be attacked than VanillaRNN.

\begin{table}[htbp]
\vspace{-.1cm}
\caption{Fooling rates in different settings on UCF101 dataset.} \vspace{-.1cm}
\centering
\begin{tabular}{|p{1.7cm}<{\centering}|p{1.7cm}<{\centering}|p{1.0cm}<{\centering}|p{1.0cm}<{\centering}|p{1.0cm}<{\centering}|}
\hline
\textbf{Models}      &VanillaRNN   &LSTM  &GRU &Pooling  \\
\hline

VanillaRNN         &$95.2\%$    &$95.2\%$      &$95.2\%$        &$71.0\%$     \\
\hline
LSTM                  &$84.1\%$    &$100\%$      &$97.1\%$       &$76.8\%$      \\
\hline
GRU                   &$81.8\%$    &$92.4\%$    &$100\%$        &$66.7\%$      \\
\hline
Pooling               &$84.1\%$    &$96.8\%$    &$95.2\%$        &$87.3\%$      \\
\hline
\end{tabular}\label{tab:tab3}
\vspace{-.4cm}
\end{table}

\subsubsection{Transferability across Videos}
We also evaluate the transferability of perturbations across videos. Th universal perturbations are computed using Eq.(\ref{eq:eq2}) on training set, and then are added to the testing videos  to evaluate their transferability. The visualization of universal perturbations can be found in Fig.~\ref{fig:figure9}. The performance (Fooling rate) is listed in Table \ref{tab:tab5}, where shows the results of our method has good transferability across videos (achieving the $95.2\%$ fooling rate on the testing set). In other words, the universal perturbations can make new videos fool the action recognition networks.
\begin{table}[htbp]
\vspace{-.1cm}
\caption{Performance of the cross-videos attack.} \centering
\vspace{-.1cm}
\begin{tabular}{|p{2.3cm}<{\centering}|p{2.0cm}<{\centering}|p{2.0cm}<{\centering}|}
\hline
\textbf{Metric}      &Training set  &Testing set   \\
\hline
Fooling rate (F)               &$100\%$            &$95.2\%$         \\
\hline
\end{tabular}\label{tab:tab5}
\vspace{-.4cm}
\end{table}

\section{Conclusions}
In this paper, we explored the adversarial perturbations for videos. An $l_{2,1}$-norm based optimization algorithm was proposed to solve this problem. The $l_{2,1}$ norm applied the $l_1$ norm across frames, and thus, can ensure the $sparsity$ of perturbations. A serial of experiments conducted on UCF101 dataset demonstrated that our method had better transferability across models and videos. More importantly,  our method showed the $propagation$ of perturbations under the $l_{2,1}$ constraint within the CNN+RNN architecture. According to this observation, we further presented the efficient method for adversarial videos based on the perturbation propagation.

\bibliographystyle{named}
\bibliography{ijcai18}

\begin{thebibliography}{}

\bibitem[\protect\citeauthoryear{Baluja and
  Fischer}{2017}]{baluja2017adversarial}
Shumeet Baluja and Ian Fischer.
\newblock Adversarial transformation networks: Learning to generate adversarial
  examples.
\newblock {\em arXiv preprint arXiv:1703.09387}, 2017.

\bibitem[\protect\citeauthoryear{Carlini and Wagner}{2017}]{carlini2017towards}
Nicholas Carlini and David Wagner.
\newblock Towards evaluating the robustness of neural networks.
\newblock In {\em IEEE Symposium on Security and Privacy (SP)}, pages 39--57,
  2017.

\bibitem[\protect\citeauthoryear{Donahue \bgroup \em et al.\egroup
  }{2017}]{donahue2015long}
Jeffrey Donahue, Lisa Anne~Hendricks, Sergio Guadarrama, Marcus Rohrbach,
  Subhashini Venugopalan, Kate Saenko, and Trevor Darrell.
\newblock Long-term recurrent convolutional networks for visual recognition and
  description.
\newblock In {\em Proceedings of the IEEE conference on computer vision and
  pattern recognition}, pages 2625--2634, 2017.

\bibitem[\protect\citeauthoryear{Dong \bgroup \em et al.\egroup
  }{2014}]{dong2014learning}
Chao Dong, Chen~Change Loy, Kaiming He, and Xiaoou Tang.
\newblock Learning a deep convolutional network for image super-resolution.
\newblock In {\em European Conference on Computer Vision}, pages 184--199.
  Springer, 2014.

\bibitem[\protect\citeauthoryear{Dong \bgroup \em et al.\egroup
  }{2017}]{dong2017boosting}
Yinpeng Dong, Fangzhou Liao, Tianyu Pang, Hang Su, Xiaolin Hu, Jianguo Li, and
  Jun Zhu.
\newblock Boosting adversarial attacks with momentum.
\newblock {\em arXiv preprint arXiv:1710.06081}, 2017.

\bibitem[\protect\citeauthoryear{Goodfellow \bgroup \em et al.\egroup
  }{2014}]{goodfellow2014explaining}
Ian~J Goodfellow, Jonathon Shlens, and Christian Szegedy.
\newblock Explaining and harnessing adversarial examples.
\newblock {\em arXiv preprint arXiv:1412.6572}, 2014.

\bibitem[\protect\citeauthoryear{He \bgroup \em et al.\egroup
  }{2016}]{he2016deep}
Kaiming He, Xiangyu Zhang, Shaoqing Ren, and Jian Sun.
\newblock Deep residual learning for image recognition.
\newblock In {\em Proceedings of the IEEE conference on computer vision and
  pattern recognition}, pages 770--778, 2016.

\bibitem[\protect\citeauthoryear{Karpathy \bgroup \em et al.\egroup
  }{2014}]{karpathy2014large}
Andrej Karpathy, George Toderici, Sanketh Shetty, Thomas Leung, Rahul
  Sukthankar, and Li~Fei-Fei.
\newblock Large-scale video classification with convolutional neural networks.
\newblock In {\em Proceedings of the IEEE conference on Computer Vision and
  Pattern Recognition}, pages 1725--1732, 2014.

\bibitem[\protect\citeauthoryear{Kingma and Ba}{2014}]{kingma2014adam}
Diederik Kingma and Jimmy Ba.
\newblock Adam: A method for stochastic optimization.
\newblock {\em arXiv preprint arXiv:1412.6980}, 2014.

\bibitem[\protect\citeauthoryear{Kurakin \bgroup \em et al.\egroup
  }{2016}]{kurakin2016adversarial}
Alexey Kurakin, Ian Goodfellow, and Samy Bengio.
\newblock Adversarial examples in the physical world.
\newblock {\em arXiv preprint arXiv:1607.02533}, 2016.

\bibitem[\protect\citeauthoryear{Liu \bgroup \em et al.\egroup
  }{2016}]{liu2016delving}
Yanpei Liu, Xinyun Chen, Chang Liu, and Dawn Song.
\newblock Delving into transferable adversarial examples and black-box attacks.
\newblock {\em arXiv preprint arXiv:1611.02770}, 2016.

\bibitem[\protect\citeauthoryear{Moosavi-Dezfooli \bgroup \em et al.\egroup
  }{2016a}]{moosavi2016universal}
Seyed-Mohsen Moosavi-Dezfooli, Alhussein Fawzi, Omar Fawzi, and Pascal
  Frossard.
\newblock Universal adversarial perturbations.
\newblock {\em arXiv preprint arXiv:1610.08401}, 2016.

\bibitem[\protect\citeauthoryear{Moosavi-Dezfooli \bgroup \em et al.\egroup
  }{2016b}]{moosavi2016deepfool}
Seyed-Mohsen Moosavi-Dezfooli, Alhussein Fawzi, and Pascal Frossard.
\newblock Deepfool: a simple and accurate method to fool deep neural networks.
\newblock In {\em Proceedings of the IEEE Conference on Computer Vision and
  Pattern Recognition}, pages 2574--2582, 2016.

\bibitem[\protect\citeauthoryear{Nguyen \bgroup \em et al.\egroup
  }{2015}]{nguyen2015deep}
Anh Nguyen, Jason Yosinski, and Jeff Clune.
\newblock Deep neural networks are easily fooled: High confidence predictions
  for unrecognizable images.
\newblock In {\em Proceedings of the IEEE Conference on Computer Vision and
  Pattern Recognition}, pages 427--436, 2015.

\bibitem[\protect\citeauthoryear{Poppe}{2010}]{poppe2010survey}
Ronald Poppe.
\newblock A survey on vision-based human action recognition.
\newblock {\em Image and vision computing}, 28(6):976--990, 2010.

\bibitem[\protect\citeauthoryear{Simonyan and
  Zisserman}{2014}]{simonyan2014two}
Karen Simonyan and Andrew Zisserman.
\newblock Two-stream convolutional networks for action recognition in videos.
\newblock In {\em Advances in neural information processing systems}, pages
  568--576, 2014.

\bibitem[\protect\citeauthoryear{Soomro \bgroup \em et al.\egroup
  }{2012}]{soomro2012ucf101}
Khurram Soomro, Amir~Roshan Zamir, and Mubarak Shah.
\newblock Ucf101: A dataset of 101 human actions classes from videos in the
  wild.
\newblock {\em arXiv preprint arXiv:1212.0402}, 2012.

\bibitem[\protect\citeauthoryear{Szegedy \bgroup \em et al.\egroup
  }{2013}]{szegedy2013intriguing}
Christian Szegedy, Wojciech Zaremba, Ilya Sutskever, Joan Bruna, Dumitru Erhan,
  Ian Goodfellow, and Rob Fergus.
\newblock Intriguing properties of neural networks.
\newblock {\em arXiv preprint arXiv:1312.6199}, 2013.

\bibitem[\protect\citeauthoryear{Wang and Yeung}{2013}]{wang2013learning}
Naiyan Wang and Dit-Yan Yeung.
\newblock Learning a deep compact image representation for visual tracking.
\newblock In {\em Advances in neural information processing systems}, pages
  809--817, 2013.

\bibitem[\protect\citeauthoryear{Wang \bgroup \em et al.\egroup
  }{2016}]{wang2016temporal}
Limin Wang, Yuanjun Xiong, Zhe Wang, Yu~Qiao, Dahua Lin, Xiaoou Tang, and Luc
  Van~Gool.
\newblock Temporal segment networks: Towards good practices for deep action
  recognition.
\newblock In {\em European Conference on Computer Vision}, pages 20--36.
  Springer, 2016.

\bibitem[\protect\citeauthoryear{Wright \bgroup \em et al.\egroup
  }{2009}]{wright2009robust}
John Wright, Allen~Y Yang, Arvind Ganesh, S~Shankar Sastry, and Yi~Ma.
\newblock Robust face recognition via sparse representation.
\newblock {\em IEEE transactions on pattern analysis and machine intelligence},
  31(2):210--227, 2009.

\bibitem[\protect\citeauthoryear{Xie \bgroup \em et al.\egroup
  }{2017}]{xie2017adversarial}
Cihang Xie, Jianyu Wang, Zhishuai Zhang, Yuyin Zhou, Lingxi Xie, and Alan
  Yuille.
\newblock Adversarial examples for semantic segmentation and object detection.
\newblock In {\em IEEE International Conference on Computer Vision}, 2017.

\bibitem[\protect\citeauthoryear{Yang \bgroup \em et al.\egroup
  }{2010}]{yang2010image}
Jianchao Yang, John Wright, Thomas~S Huang, and Yi~Ma.
\newblock Image super-resolution via sparse representation.
\newblock {\em IEEE transactions on image processing}, 19(11):2861--2873, 2010.

\bibitem[\protect\citeauthoryear{Yue-Hei~Ng \bgroup \em et al.\egroup
  }{2015}]{yue2015beyond}
Joe Yue-Hei~Ng, Matthew Hausknecht, Sudheendra Vijayanarasimhan, Oriol Vinyals,
  Rajat Monga, and George Toderici.
\newblock Beyond short snippets: Deep networks for video classification.
\newblock In {\em Proceedings of the IEEE conference on computer vision and
  pattern recognition}, pages 4694--4702, 2015.

\end{thebibliography}

\end{document}